\pdfoutput=1

\documentclass[11pt,a4paper]{article}
\usepackage{authblk}
\usepackage[hyperref]{emnlp2021}
\usepackage{times}
\usepackage{latexsym}

\usepackage{graphicx}
\usepackage{tabularx}
\usepackage{amsmath}
\usepackage{float}
\usepackage{microtype}
\usepackage{multirow}
\usepackage{makecell}
\usepackage{stfloats}
\usepackage{color}

\usepackage{microtype}

\title{An unsupervised framework for tracing textual sources of moral change}

\author{Aida Ramezani$^{1*}$, Zining Zhu$^{1,2*}$, Frank Rudzicz$^{1,2,3}$, Yang Xu$^{1,2,4}$ \\
$^1$ Department of Computer Science, University of Toronto, Toronto, Canada \\
$^2$ Vector Institute for Artificial Intelligence, Toronto, Canada \\ 
$^3$ St. Michael's Hospital, Toronto, Canada \\
$^4$ Cognitive Science Program, University of Toronto, Toronto, Canada \\
\texttt{\{armzn, zining, frank, yangxu\}@cs.toronto.edu} 
}

\date{}

\begin{document}
\maketitle
\begin{abstract}
Morality plays an important role in social well-being, but people's moral perception is not stable and changes over time. Recent advances in natural language processing have shown that text is an effective medium for informing moral change, but no attempt has been made to quantify the origins of these changes. We present a novel unsupervised framework for tracing textual sources of moral change toward entities through time. We characterize moral change with probabilistic topical distributions and infer the source text that exerts prominent influence on the moral time course. We evaluate our framework on a diverse set of data ranging from social media to news articles. We show that our framework not only captures fine-grained human moral judgments, but also identifies coherent source topics of moral change triggered by historical events. We apply our methodology to analyze the news in the COVID-19 pandemic and demonstrate its utility in identifying sources of moral change in high-impact and real-time social events.
\end{abstract}



\section{Introduction}
\label{sec:intro}

From ancient Greek scholars to philosophers of the past centuries, morality has been a subject of central importance in human history \citep{plato, aristotle,hume1739treatise, smiththeory, kantgroundwork,nietzsche1900genealogie}. Despite this importance, people's morals are not static but change over time~\citep{bloom2010morals}. Recent advances in natural language processing (NLP) have shown that text can inform moral sentiment and its change over time (e.g., how slavery was increasingly perceived to be morally wrong)~\cite{xie2020text,garten2016morality}. However, critically under-explored are the origins of these changes. We present a framework for tracing textual sources of moral change that requires minimal human intervention or supervision.

The study of moral sentiment is a prominent subject in social psychology \citep{piaget1932moral, kohlberg1969stage, kohlberg1977moral, haidt2001emotional, pizarro2003intelligence}, and the advent of Moral Foundations Theory \citep{graham2013moral} has provided an impetus for text-based analysis of moral sentiment in natural language processing. 
Existing studies range from moral sentiment  classification to temporal inference of moral sentiment change (e.g., \citealp{garten2016morality, mooijman2018moralization, lin2018acquiring, xie2020contextualized,xie2020text}). 

The problem we focus on here is how moral perception toward entities (e.g., political leaders) varies through time, and whether textual analysis can help extract the sources of this variation. For instance, an entity like \emph{Bill Clinton} could be applauded for charity at one time but deprecated for a sex scandal at another time. Similarly, moral sentiment toward a more general entity like \emph{policemen} could undergo a negative shift  due to acts on racial discrimination. Existing methods for moral sentiment detection typically take an aggregate approach and do not focus on analyzing moral sentiment of entities \citep{garten2016morality,lin2018acquiring,mooijman2018moralization,xie2020text}. Here, we develop a methodology to identify textual sources that give rise to moral sentiment change toward an entity. Our work takes a similar approach to detecting sources of gender bias in text by locating a set of documents that influence gender bias in word embeddings \citep{brunet2019understanding}.

We propose a probabilistic unsupervised framework informed by both textual inference of moral sentiment and dynamic topic model~\citep{blei2006dynamic}. Capturing events as topic distributions, we approach this problem by decomposing textual mentions of an entity into topics and quantifying the contributions of different topics toward moral sentiment of an entity. We attribute the origins of moral change as topics that contribute saliently to changes in the time course of moral sentiment. We compare this  approach with classic work on influence function \cite{cook1980characterizations}, which has been used to quantify the effect of samples in statistical estimation (see also \citealp{koh2017understanding,brunet2019understanding}).

\begin{figure}[t!]
\centering
\includegraphics[width=0.47\textwidth]{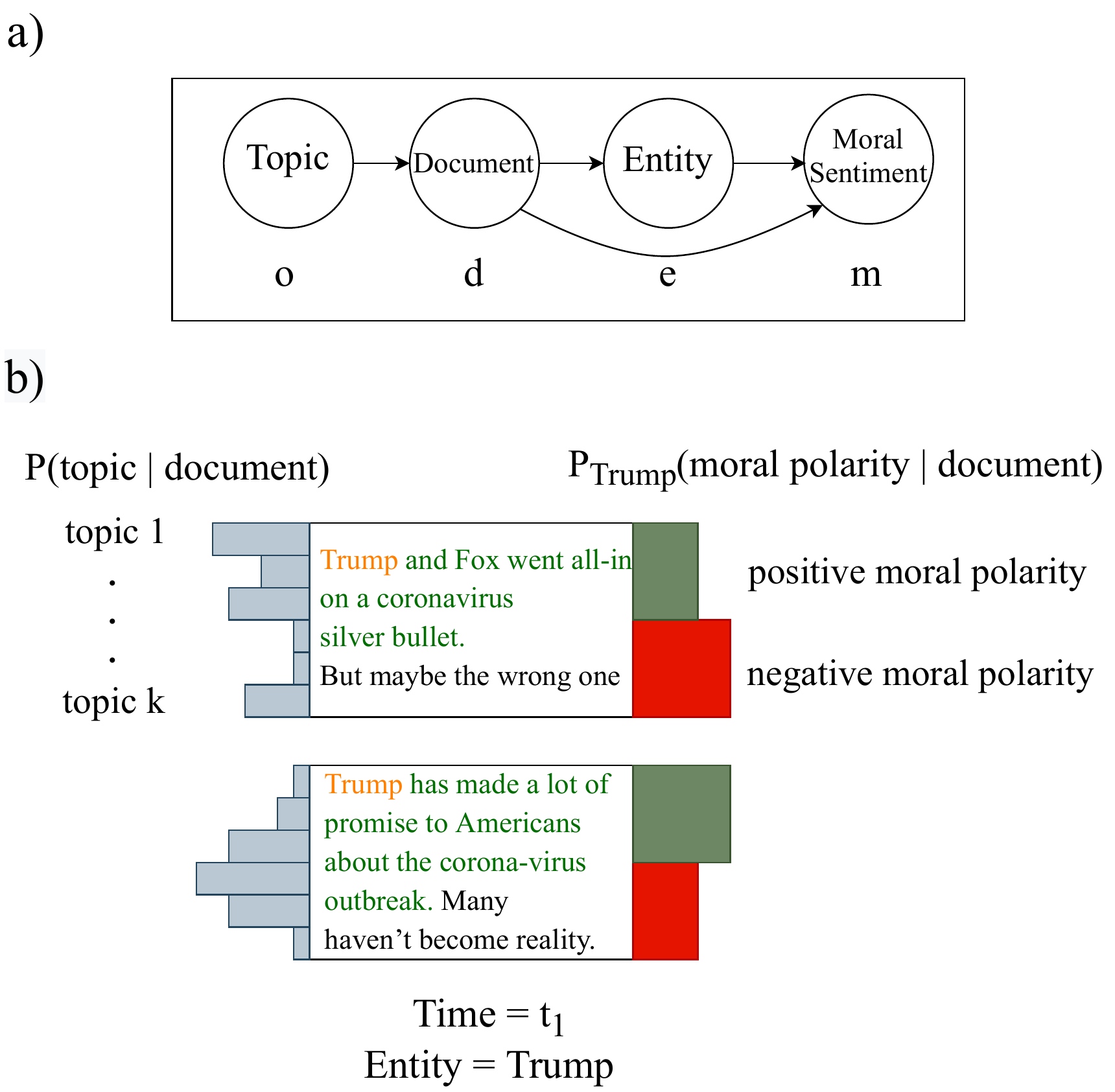}
\caption{Graphical model and illustration of our framework for topic-based source analysis of moral change.}
\label{fig:general_method}
\end{figure}

Figure~\ref{fig:general_method} illustrates our framework. Figure~\ref{fig:general_method}a shows the generative process in our topic-based approach. Given an entity (e.g., {\it Donald Trump}) and its mentions in a set of documents (e.g., tweets or news articles), we wish to infer the most salient source topic(s) that gave rise to changes in the moral sentiment time course about that entity. Here as an illustration the moral sentiment toward {\it Donald Trump} is analyzed through a set of news articles. Each article includes mentions of this entity, specified as topical distributions that contribute toward the perceived moral sentiment of the entity (see Figure~\ref{fig:general_method}b). As moral sentiment of the entity changes over time, our framework uses probabilistic inference jointly with change point analysis to extract the most salient topic and its relevant source documents that underlie these changes. We show how our approach predicts fine-grained human moral judgment variation across topics and identifies influential and coherent text as the sources of moral change for both historical and modern events.

\section{Related work on textual inference of moral sentiment in NLP}
\label{sec:rw}
The development of Moral Foundations Theory (MFT) jointly with Moral Foundations Dictionary (MFD)  \citep{graham2009liberals,graham2013moral} has propelled recent research in the natural language processing community to explore automated textual inference of moral sentiment. MFT sought to explain the cultural variation in morality and moral concerns along five or six moral foundations, each organized in terms of the polarities virtue (+) and vice ($-$).

The computational methods using MFT tend to rely on supervised approaches to predicting the moral sentiment reflected in text \citep{garten2016morality, lin2018acquiring, mooijman2018moralization,xie2020contextualized}. Other related work has  characterized moral biases  in language models \citep{schramowski2019bert, jentzsch2019moral, xie2020text}, and contributed new datasets for tasks such as automatic ethical judgment and inference of sociomoral norms \citep{hoover2020moral, lourie2020scruples, forbes2020social}. Existing work  has also studied moral sentiment change over time \citep{xie2020text} showing how word embeddings  capture hidden moral biases underlying different concepts (e.g., \emph{slavery}) in history. This model uses MFD words as seeds and a hierarchical framework to capture moral change in three tiers: moral relevance, moral polarity, and fine-grained moral foundations.\footnote{The 10 foundation categories follow from the Moral Foundations Theory including 5 opposing pairs: Care(+)/Harm($-$),\ Fairness(+)/Cheating($-$),\ Loyalty(+)/Betrayal($-$),\ Authority(+)/Subversion($-$),\ and Sanctity(+)/Degradation($-$).} 

Here we go beyond this line of research by developing an unsupervised framework that automatically identifies sources of moral change toward entities in text.

\section{Methodology}

\label{sec:methodology}

We formulate textual source tracing of moral change as a probabilistic inference problem.  This model  allows us to identify the sources of change at the topic level (source topic), as well as retrieve a  set of related documents (source documents) underlying the detected moral  change. To do so, we need to quantify 1) the moral time course of an entity based on textual input, and 2) the influence of topics on the changes in moral time course.\footnote{Codes to replicate the analyses are publicly available at \url{https://github.com/AidaRamezani/moral-source-tracing}}

{\bf Quantification of moral time course.} We estimate the moral sentiment along moral dimension $m$ for entity $e$ at time point $t$ as follows: 

\begin{equation}
\begin{aligned}
         P(m | e, t) = \frac{
    \sum_{d \in D_{e,t}} P_{e}(m| d)} {|D_{e,t}|}
\end{aligned}
\label{eq:1}
\end{equation}


Here $D_{e,t}$ is the set of documents (indexed by $d$) at time point $t$ that contain entity $e$ at least once.\footnote{We use the co-reference resolution module \texttt{neuralcoref} implemented in \texttt{spaCy} to find all the mentions of an entity in a document. We describe the details of the pre-processing in Appendix A.} For example, $D_{e,t}$ can be all the documents in our corpus that are published in $t = $ December 1997, and include a mention of entity $e = $ {\it Bill Clinton}. Moral dimension $m$ can be  moral relevance, moral polarity, or one of the moral foundations in MFT. 

To construct a vector representation for a document, we exclude all the sentences in $d$ that do not include any mentions of entity $e$. After lemmatizing the rest of the document using  \texttt{spaCy} English model, we remove 1) function words, 2) entity $e$ and its mentions, and 3) words that are classified as morally irrelevant by the centroid model following \citet{xie2020text}. We then derive the vector representation of document $d$ by taking an average of the semantic vector representations (i.e., word embeddings) of the remaining words: $V_d = \frac{1}{|d|} \sum_{w \in d} V_w$. Here $V_d$ is the vector representation of document $d$, and $V_w$ is the vector representation of  lemma $w$. 

To estimate $P_{e}(m | d)$, we use the centroid model in~\citet{xie2020text}. This model estimates this probability by comparing the similarity of an input vector (i.e., $V_d$) to each of its centroids. The centroids of this model in the moral relevance tier are the average word embeddings of MFD words and a set of morally neutral words. For moral polarity, the centroids are based on moral virtue and vice words from MFD. For the fine-grained tier, there are 10 centroids, each being the average word embeddings of the words in a moral foundation.

{\bf Quantification of textual source and influence of moral change.} To quantify sources of moral change, we first use the dynamic topic model \citep{blei2006dynamic} to infer emerging topics based on the temporal collection of documents that contain entity $e$ (illustrated in Figure~\ref{fig:general_method}). 
Using a dynamic topic model offers the flexibility to update old and dated topics with emerging topics over time. For entity $e$ with $k$ associated topics, we then define metric $\Delta{S}$ to quantify the influence of each topic on moral change toward this entity in time window $t \sim t+\Delta{t}$ (excluding $t$). Similar to Equation~\ref{eq:1}, $t$ is a point in time, e.g., December 1997, and $\Delta{t}$ is a time period, e.g., 3 months. Formally, this metric is as follows:
\begin{equation}
    \begin{aligned}
     &\Delta{S}(e, m, o, t, \Delta{t}) =\\
    | P(m | e , t\sim\;&t+\Delta{t},topic\neq o) -
     P(m | e , t) | 
    \end{aligned}
    \label{eq:2}
\end{equation}
Here $o$ represents a  topic ranging from $1$ to $k$, and $\Delta{S}$ measures the degree to which removing a topic can restore the moral sentiment to its base state. The topic with the lowest $\Delta{S}$ is  the most influential source for the change. We derive $P(m | e , t\sim t+\Delta{t},  topic \neq o)$ as follows:

\begin{equation}
\begin{aligned}
P(m | &e , t \sim t+\Delta{t},  topic \neq o)  \\
 &= \sum_{d \in D_{e, t\sim  t+\Delta{t}}}  P_{e} (m | d) P(d | topic \neq o) \\
&\propto \sum_{d \in D_{e, t\sim t+\Delta{t}}}  P_{e} (m | d) P(topic \neq o | d) \\
&\propto \sum_{d \in D_{e, t\sim  t+\Delta{t}}} P_{e} (m | d) (1 - P(topic = o | d))
\end{aligned}
\label{eq:3}
\end{equation}

We estimate $P(topic = o | d)$ from the dynamic topic model. Similar to Equation~\ref{eq:1}, $D_{e, t\sim t+\Delta{t}}$ represents the documents that contain entity $e$ appearing within time window $t \sim t+\Delta{t}$. Without loss of generality, we assume a uniform prior for the distribution of the documents, so $P(d)$ is constant.

We detect significant changes in moral time course using an established method for change-point detection \cite{kulkarni2015statistically}. Given a time series as the input, this method first generates random perturbations of the time series and compares the magnitude of the mean shift before and after a time point in the original series to that in the random perturbations, for all the points in the time course individually. The outputs of the algorithm will be the time points with the most significant mean shifts (i.e., lowest p-values) as the change points. We consider a sliding window with a size of $W_t$ time points and a step size of $W_s$ and run the change point detection algorithm on the moral sentiment time series of an entity by estimating the probability in Equation~\ref{eq:1} incrementally over time. This gives change point(s) $t$ and the relevant time window(s) $\Delta{t}$ which we use in Equations~\ref{eq:2} and~\ref{eq:4}. We find  $W_t = 7$ and $W_s = 3$ to be reasonable choices.

We define $\Delta{J}$ to quantify the
degree of influence of a set of documents $D^*$ (appearing at time $t \sim t+\Delta{t}$) on moral change toward entity $e$ in moral dimension $m$ at time $t$. We compute this by calculating how the entity-based moral  change is impacted by the removal of $D^*$,  formally as:

\begin{equation}
    \begin{aligned}
    \Delta{J}(e, m, D^*, t, \Delta{t}) = & \\
    | P(m | e , t \sim t+\Delta{t}, D_{e, t \sim t+\Delta{t}} \setminus D^*) - &
    P(m | e , t) |
    \end{aligned}
    \label{eq:4}
\end{equation}

Here $P(m | e , t \sim t+\Delta{t}, D_{e, t \sim t+\Delta{t}} \setminus D^*)$ is calculated using  Equation~\ref{eq:1} over the documents including entity $e$ appearing at $t \sim t + \Delta{t}$ excluding set $D^*$. The difference between $\Delta{S}$ and $\Delta{J}$ is that $\Delta{S}$ measures the influence of a topic over all the documents in a probabilistic setting, whereas $\Delta{J}$ measures the influence of a set of documents regardless of their  topic associations.

\section{Experiments and results}

We evaluate and apply our framework in three diverse and real-world settings.

\subsection{Datasets}

{\bf Moral Foundations Twitter Corpus (MFTC).} We use Moral Foundations Twitter Corpus \citep{hoover2020moral} for the first case study. This corpus provides a large set of human judgments along different moral dimensions for tweets divided into distinct topics. Each tweet is hand-annotated for the 10 foundation categories and moral relevance. Using Twitter Developer Account, we were able to extract 21,482 tweets  falling under six topic domains specified in the original dataset: ALM (all lives matter), BLM (black lives matter), Baltimore, Davidson, Election, and Sandy. 

{\bf New York Times Annotated Corpus (NYT).} We use the New York Times Annotated Corpus \citep{nyt} for the second case study. This dataset contains over 1.8 million news articles published in the New York Times from 1987 to 2007.  

{\bf COVID-19 News Dataset (COVID).} We use AYLIEN Free Coronavirus Dataset in the third case study.\footnote{\url{https://aylien.com/blog/free-coronavirus-news-dataset}} This dataset contains more than 1,500,000 annotated English news articles relevant to the COVID-19 pandemic. We include the articles published in well-known United States news agencies from January, 2020 to the end of July 2020. We extracted a total number of 94,732 articles from CNN, Foxnews, NBC News, The New York Times, USA Today, abc News, CBS News, Washington Post, MSNBC News, and Los Angeles Times.

\subsection{Evaluation on human moral judgment} 
\label{sec:mftc}

Human moral sentiment toward entities may vary across topical contexts. As an initial study, we show how this variation is present in social media and can be captured by a topic-based approach where topic information is given. We use the MFTC tweet data for evaluation, based on the moral judgment of tweets in 6 topics: ALM, BLM, Baltimore, Davidson, Election, and Sandy. We summarize the human moral judgment of entities across topics using a count-based measure. Specifically, we compute the empirical probability $\widehat{P}(m | e, o)$ for moral dimension $m$, entity $e$, and topic $o$ as $\widehat{P}(m | e, o) = \frac{count(m, e, o)}{count(e, o)}$. Here $count(e, o)$ is the number of tweets in topic $o$ that contain entity $e$. To calculate $count(m, e, o)$, we count the number of tweets from topic $o$ that contain entity $e$, and were annotated with moral sentiment dimension $m$ in MFTC. To prepare ground-truth data, we take the following steps: 1) For the moral relevance dimension, if more than half of the annotators annotate a tweet ``non-moral'', we consider the tweet as morally irrelevant. 2) For the moral polarity dimension, if the majority of annotations fall under the positive fine-grained categories, the moral polarity of the tweet is positive (and negative otherwise). 3) For the foundation categories, each tweet is given the label of the category receiving the majority vote from the annotators. If more than one category satisfies this condition, we randomly assign one of them to the tweet. We also used graded proportions instead of binary ground-truth labels and obtained similar results. We analyzed moral judgment on the  $53$ most frequent entities in the MFTC that appear under at least two topics. The entities include hashtags, mentions, and the named entities such as people, organizations, groups, and concepts.\footnote{The named entities are detected using the NER in \texttt{spaCy}. We manually check these entities, and map all the forms of an entity to a single unique type (e.g., Barack Obama and Obama are both considered the same entity).}


\begin{table*}[ht]
\centering
\resizebox{\linewidth}{!}{
\begin{tabular}{lccc|ccc|ccc}
 \multirow{2}{*}{\makecell{\textbf{Moral Foundation}}}&
    \multicolumn{3}{c}{\makecell{Topic-based Model \\(Static Embedding)}}&
    \multicolumn{3}{c}{\makecell{Topic-free Model \\(Static Embedding)}}&
    \multicolumn{3}{c}{\makecell{Topic-free Model \\(Contextual Embedding)}}\\
    \cline{2-10}
{} & $F_1$ & Pearson's r & n & $F_1$ &Pearson's r & n & $F_1$ & Pearson's r & n
\\
\Xhline{2pt}
Moral Relevance	& \textbf{1}& \textbf{0.307} & 195  & \textbf{1} & 
 \hskip 2mm 0.098$^{-}$ & 195  & \textbf{1} & \hskip 2mm 0.103$^{-}$ & 195 \\
\hline 
Moral Polarity & \textbf{0.947} &\textbf{0.808} & 171 & \textbf{0.947} & 0.638 & 171   & 0.841 &  0.763	& 127 \\
\hline
Authority & \textbf{0.924} & 0.285 & 157 & 0.689  &\hskip 2mm 0.199$^{-}$ & 157 &   0.699 & \textbf{0.305} & 94 \\
Subversion & \textbf{0.877} & \textbf{0.251} & 143 & 0.705 & \hskip 2mm $-$0.028$^{-}$ & 143 & 0.777 & \hskip 2mm 0.242$^{-}$ & 110 \\
Care &\textbf{0.924} & \textbf{0.500} &157 & 0.689 & 0.328 & 157 & 0.699 & 0.451 & 94 \\
Harm &\textbf{0.877} & \hskip 2mm 0.060$^{-}$ &143 & 0.705 & \hskip 2mm 0.036$^{-}$ & 143 & 0.777 & \textbf{0.286} & 110 \\
Fairness & \textbf{0.924}  &\textbf{0.587} & 157 & 0.689 & 0.391 & 157 & 0.699 & 0.551 & 94 \\
Cheating & \textbf{0.877} & \textbf{0.341} & 143 & 0.705 & \hskip 2mm 0.193$^{-}$ & 143 & 0.777 & \hskip 2mm 0.125$^{-}$ & 110 \\
Loyalty &  \textbf{0.924} & \textbf{0.634} & 157 & 0.689 & 0.524 & 157 & 0.699 & \hskip 2mm 0.236$^{-}$ & 94  \\
Betrayal & \textbf{0.877} & \hskip 2mm 0.125$^{-}$ & 143 & 0.705 & \hskip 2mm 0.045$^{-}$ & 143 & 0.777 & \hskip 2mm $-$0.104$^{-}$ & 110\\
Sanctity & \textbf{0.924} & \textbf{0.526} &157 &  0.689 & 0.354 & 157  & 0.699 & 0.366 & 94 \\
Degradation & \textbf{0.877} & 0.386 & 143 & 0.705 & 0.434 & 143 & 0.777 & \textbf{0.524} & 110 \\
\end{tabular}}
\caption{Evaluation of topic-based and topic-free models in predicting fine-grained human moral judgments, based on both F$_{1}$ score and Pearson's correlation. Superscript minus sign under ``Pearson's r'' indicates $p > 0.05$ (Bonferroni corrected).}
\label{tab:mftc}
\end{table*}

We first consider a topic-based model that explicitly uses topic information and applies static word embeddings to infer moral sentiment variation across topics. For each moral dimension $m$, entity $e$, and topic $o$ we derive the following probability using the methodology from Section~\ref{sec:methodology}: $P(m | e, o) \propto \sum_{d \in tweets} P_{e}(m | d)  P(o | d)$. We use Word2Vec word embeddings \citep{mikolov2013distributed} to represent each tweet as a single vector. We also consider two alternative topic-free models using static and contextual embeddings, where topic information is discarded in moral sentiment inference, i.e., $P(m | e, o)=P(m | e)$. We use BERT \citep{devlin2018bert} in the contextual embedding model to represent tweets. Similarly for the centroid model, instead of using the static embeddings of the seed words in MFD, we use BERT to embed their definitions from the online version of the Oxford English Dictionary (OED).\footnote{\url{https://www.oed.com}}

Each model infers $P(m | e, o)$ for all entities, topics, and moral dimensions. We compare these probabilities with ground-truth moral judgments $\widehat{P}(m | e , o)$ using both F$_{1}$ score and Pearson's correlation. We consider estimates of $\widehat{P}(m | e, o)$ and $P(m | e, o)$ meaningful if 1) the entity appears in at least one of the tweets in topic $o$, and 2) there is at least one tweet in topic $o$ containing entity $e$ that satisfies the 3-tier hierarchical structure, i.e., moral polarity of an entity is only estimated when it is morally relevant, and virtuous/vice moral foundations sentiments are estimated only for morally positive/negative input. A correlation test is performed on the samples that satisfy the two criteria in both model and human judgment. F$_{1}$ score quantifies the proportion of samples that they agree on. 

Table~\ref{tab:mftc} summarizes our results in this task. We observe that the topic-based  model best accounts for the variation in human moral judgment across topics for the entities analyzed, both in terms of the F$_{1}$ scores and fine-grained correlation values. For example, the entity \emph{USA} bears an overall negative moral polarity, while the same entity appears more morally positive in tweets concerning the topic \emph{Election}. Another example is that the entity \emph{CNN} displays a negative moral polarity across all topics, but shifts to a morally positive sentiment under the topic \emph{ALM}. Our topic-based  model with static embedding captures both of these variations. These initial results provide strong support to our presumption that moral sentiment toward entities may vary across context. We next apply our framework to diachronic data where neither topic information nor change point is provided.

\subsection{Evaluation on moral source identification from news of historical events}
\label{sec:nyt}

In the second case study, we use the NYT dataset to evaluate the topic-based source model
against prominent historical events in the United States from the 20\textsuperscript{th} and 21\textsuperscript{st} centuries, and analyze the entities associated with each event. We assess the topic-based source model on its ability to identify the moral changes at the historical incidents and locate topics and source text (i.e., news articles) relevant to these events. We also use the established influence function~\cite{cook1980characterizations} as a baseline model for comparison.

\begin{figure*}[ht!]
\centering
\includegraphics[width=1\textwidth]{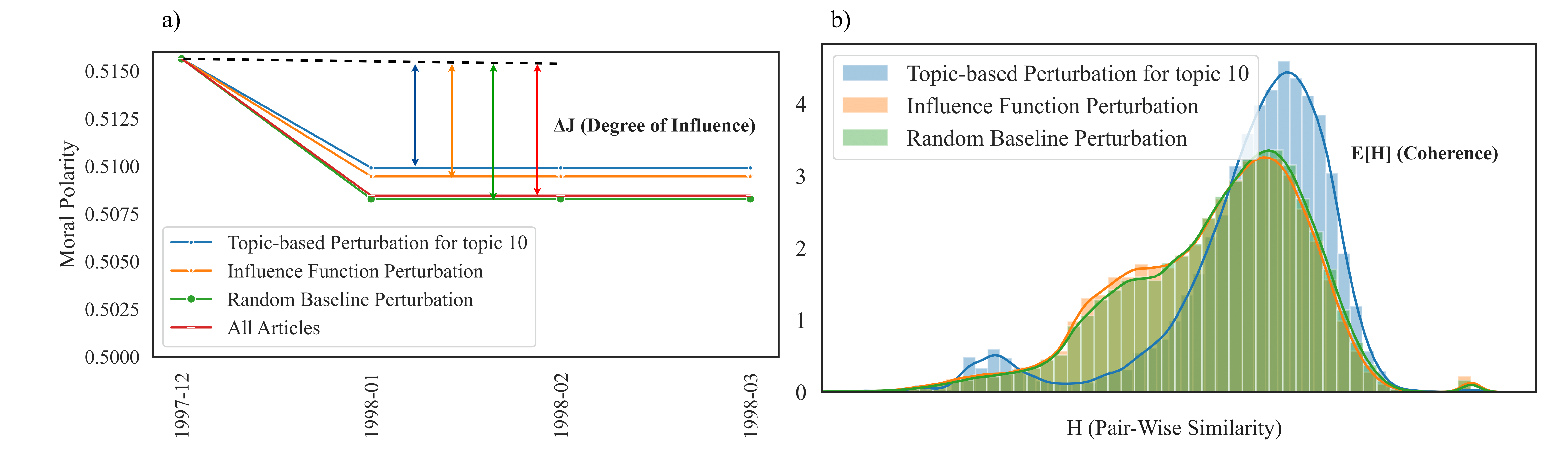}
\caption{Illustration of metrics quantifying the effectiveness of models for tracing sources of moral sentiment change based on NYT news about {\it Bill Clinton} from 1997-12 to 1998-03. a) Comparing the degree of influence ($\Delta{J}$) of different methods on restoring the change to its baseline (horizontal dash): smaller $\Delta{J}$ indicates greater influence. b) Expected coherence of source documents ($E[H]$) of different methods.}

\label{fig:method_vis}
\end{figure*}


\begin{table*}[ht]
    \centering
    \begin{tabular}{p{0.12\textwidth}|p{0.07\textwidth}|p{0.08\textwidth}|p{0.04\textwidth}|p{0.26\textwidth}|p{0.26\textwidth}}
         \textbf{\hspace{.5em} Entity}& \textbf{Initial point} & \textbf{Ending point} &\textbf{N} & \textbf{\hspace{.5em} Influence comparison} & \textbf{\hspace{.5em} Coherence comparison}  \\
         \hline
         \raisebox{0.8\height}{\makecell{George \\H. W. Bush} $\downarrow$} & \raisebox{1.3\height}{1990-07} & \raisebox{1.3\height}{1991-02} & \raisebox{1.3\height}{2829} &  \includegraphics[width=0.3\textwidth]{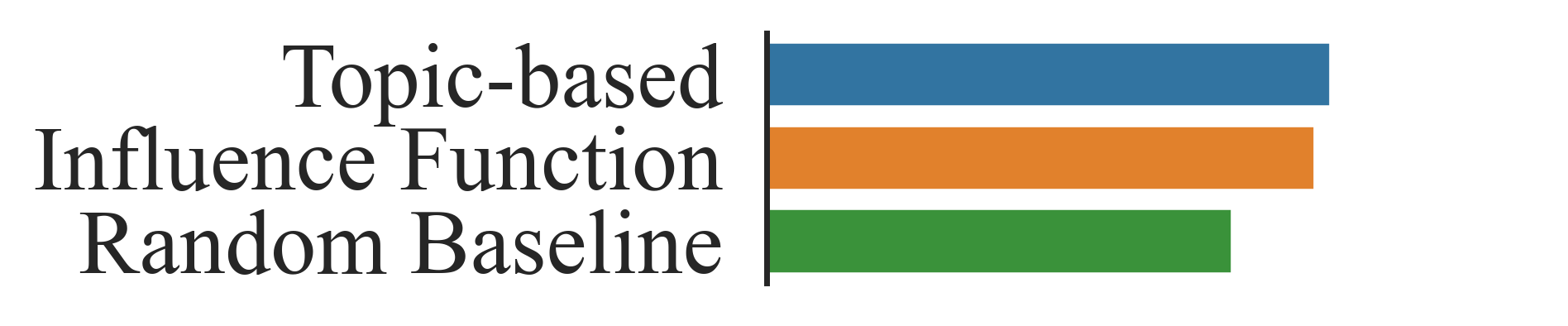} & \includegraphics[width=0.3\textwidth]{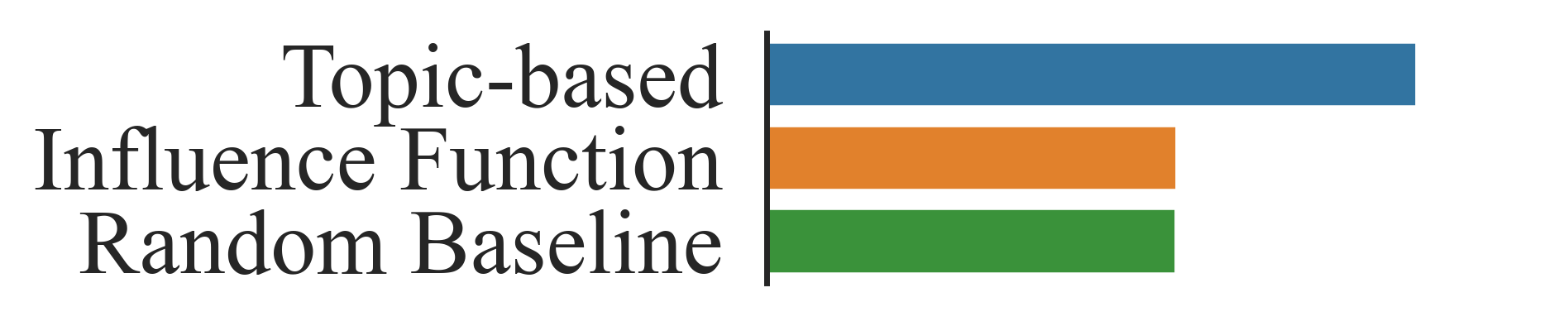}\\
         \hline
         \multicolumn{6}{c}{\makecell{The most salient topic words: iraqi, iraq, kuwait, allied, ground, saddam hussein}}
         \\
        \hline
         \raisebox{1.3\height}{\makecell{Bill Clinton $\downarrow$}} & \raisebox{1.3\height}{1997-12} & \raisebox{1.3\height}{1998-03} &\raisebox{1.3\height}{1224} &  \includegraphics[width=0.3\textwidth]{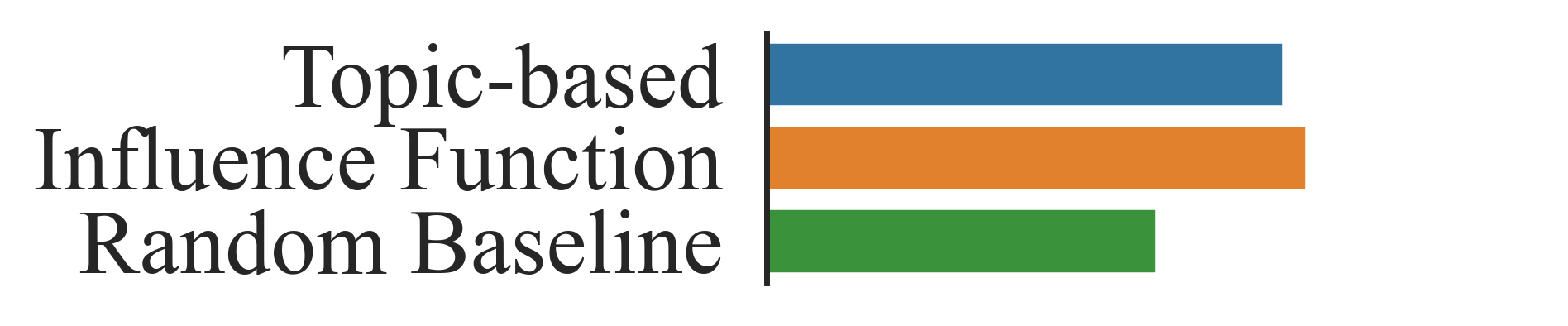} & \includegraphics[width=0.3\textwidth]{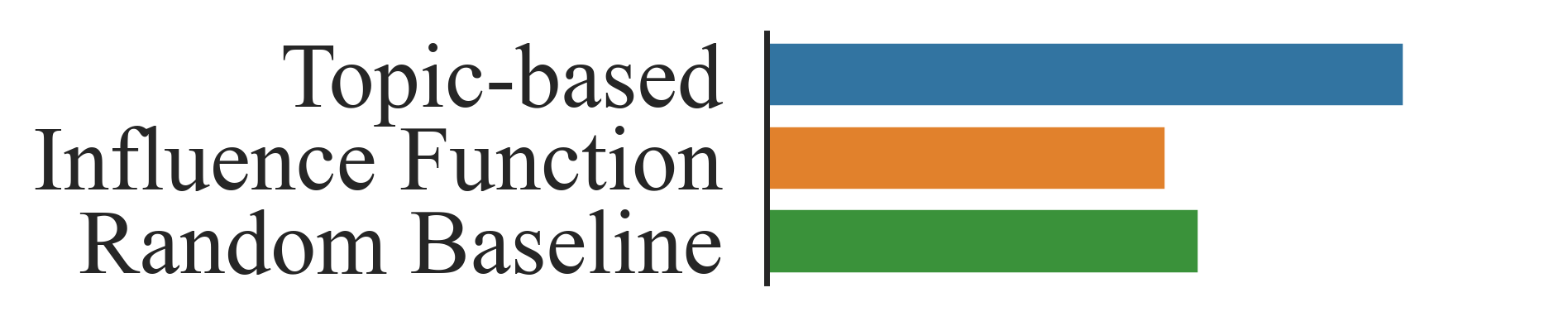}\\
         \hline
        \multicolumn{6}{c}{\makecell{The most salient topic words: intern, willey, lawyer, starr, lewinsky, babbitt, ginsburg, accusation
        }}
         \\ \hline
         \raisebox{1.3\height}{\makecell{Bill Clinton $\downarrow$}} & \raisebox{1.3\height}{1998-07} & \raisebox{1.3\height}{1998-12} & \raisebox{1.3\height}{2693} &  \includegraphics[width=0.3\textwidth]{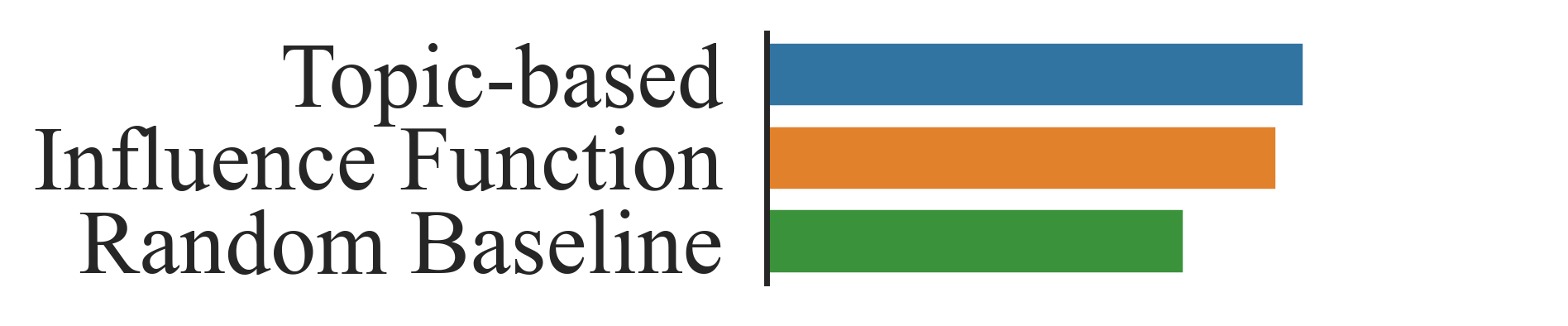} & \includegraphics[width=0.3\textwidth]{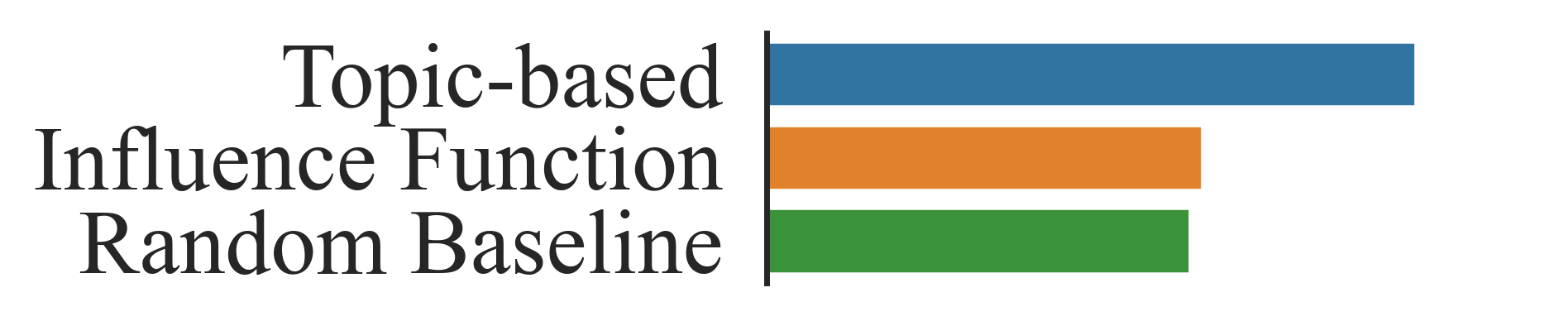}\\
         \hline
        \multicolumn{6}{c}{\makecell{The most salient topic words: censure, impeachment, impeach, judiciary, hyde, perjury}}
         \\ \hline
     \raisebox{0.8\height}{\makecell{\hspace{.25em} George\\ \hspace{.5em} Bush} $\downarrow$} & \raisebox{1.3\height}{2001-08} & \raisebox{1.3\height}{2001-12} & \raisebox{1.3\height}{2058}&  \includegraphics[width=0.3\textwidth]{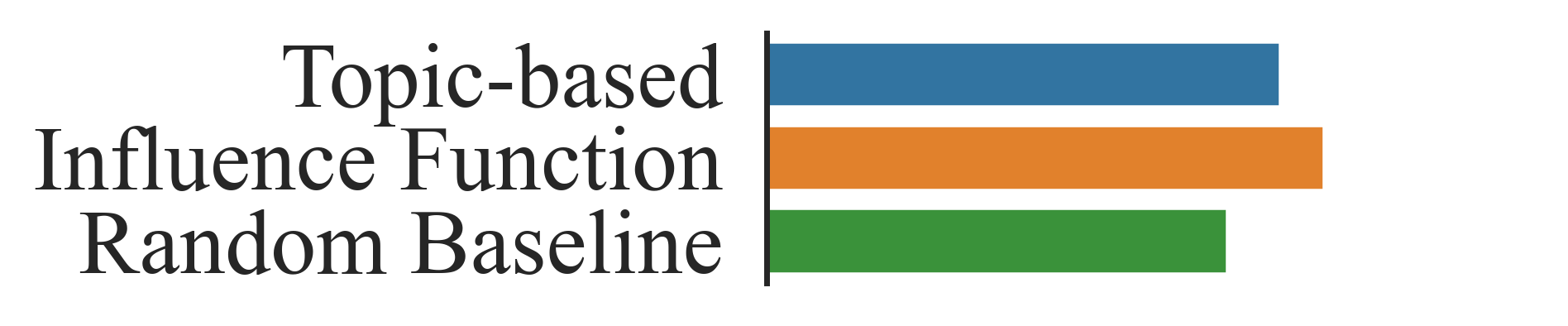} &  \includegraphics[width=0.3\textwidth]{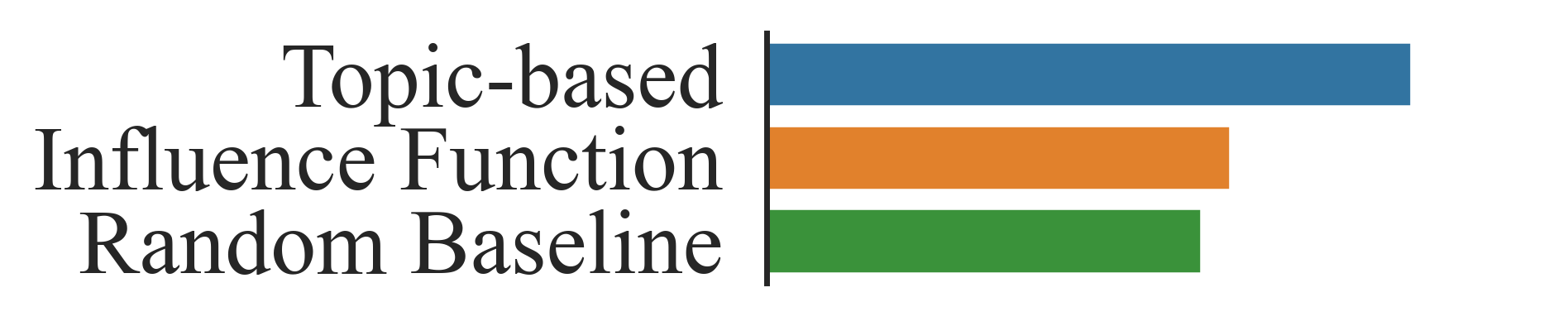}\\
         \hline
        \multicolumn{6}{c}{\makecell{The most salient topic words: al qaeda, taliban, bin laden, attack, afghan, hijacker}}
         \\ \hline
    \raisebox{1.3\height}{\makecell{China $\downarrow$}} & \raisebox{1.3\height}{2003-02} & \raisebox{1.3\height}{2003-05} & \raisebox{1.3\height}{741}& \includegraphics[width=0.3\textwidth]{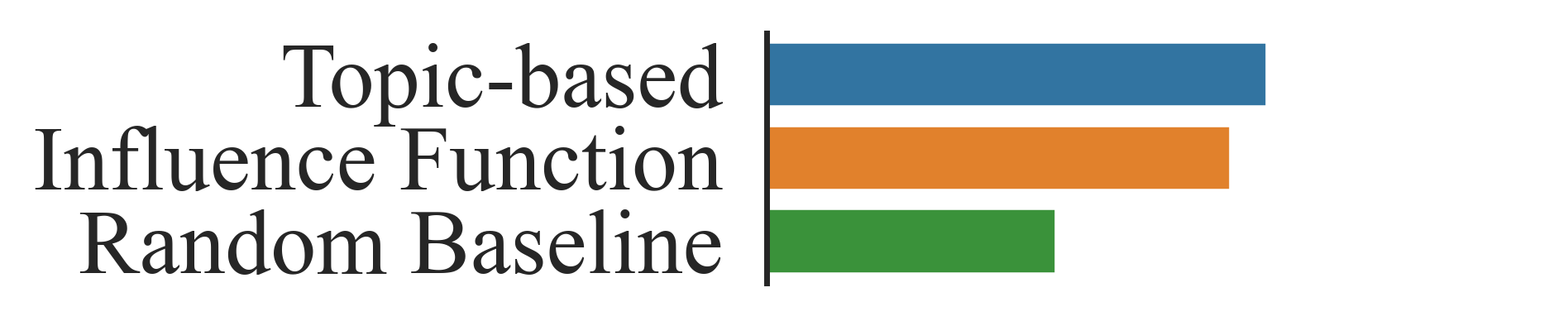} & \includegraphics[width=0.3\textwidth]{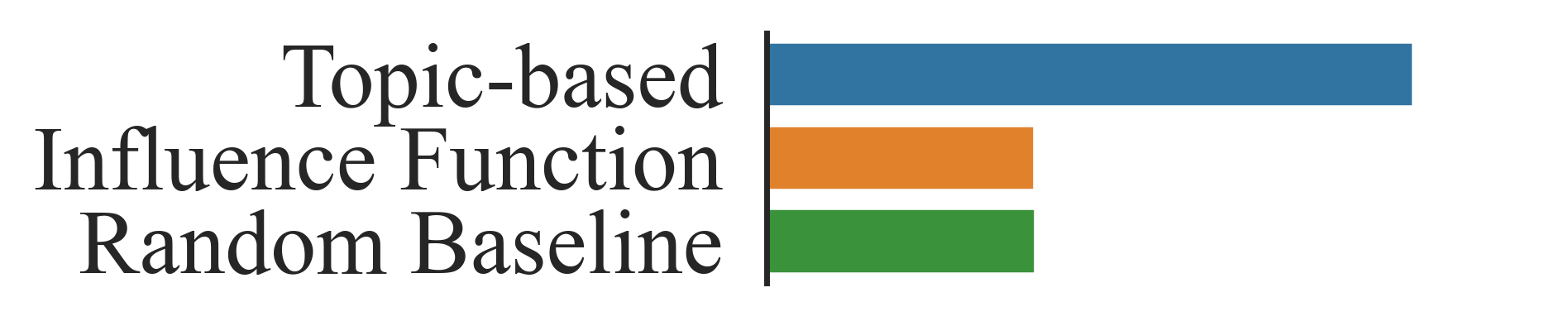}\\
         \hline
        \multicolumn{6}{c}{\makecell{The most salient topic words: sars, disease, respiratory, health, sar, outbreak, syndrome, hospital}}
         \\ \hline
        \raisebox{0.8\height}{\makecell{Saddam\\ Hussein} $\downarrow$} & \raisebox{1.3\height}{2003-04} & \raisebox{1.3\height}{2003-12} &\raisebox{1.3\height}{1546}& \includegraphics[width=0.3\textwidth]{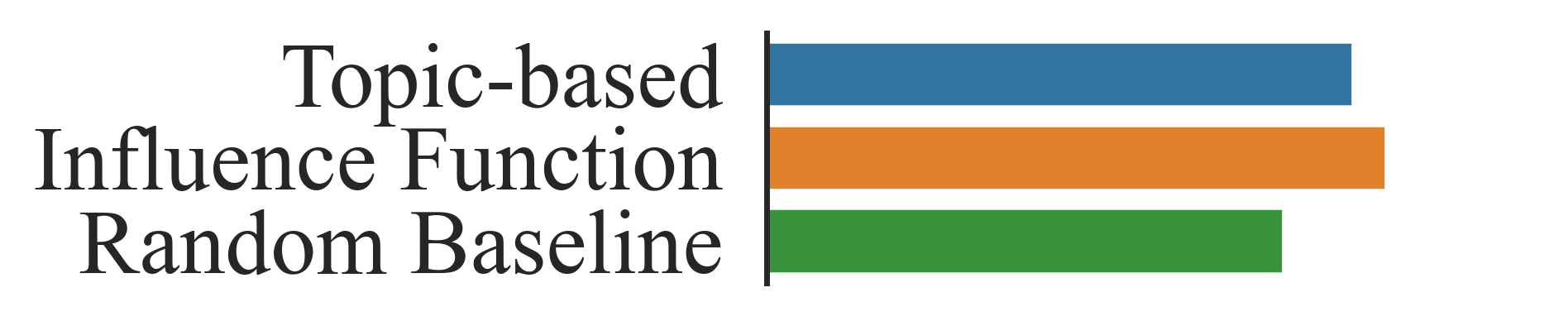} & \includegraphics[width=0.3\textwidth]{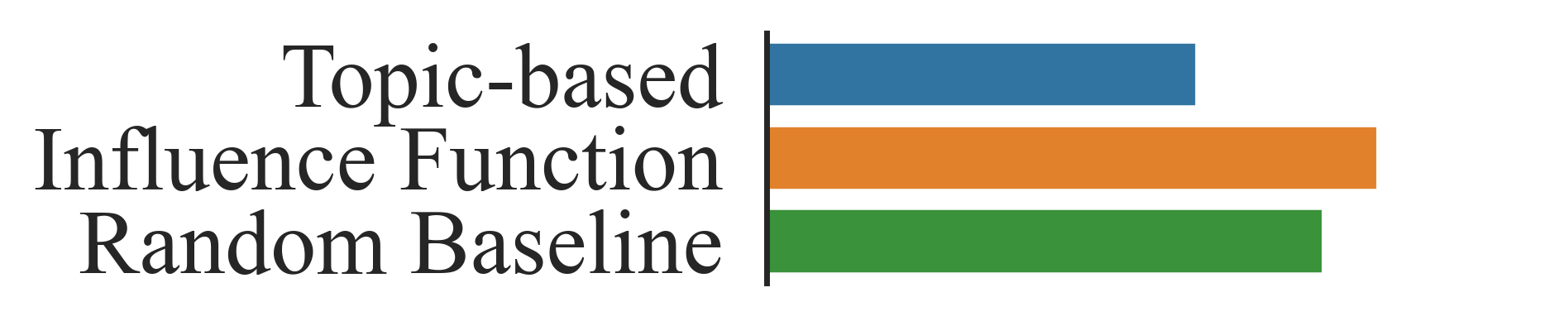}\\
         \hline
        \multicolumn{6}{c}{\makecell{The most salient topic words: dean, lieberman, kerry, howard, clark, nomination, gore}}
         \\ \hline
           \raisebox{0.8\height}{\makecell{\hspace{.25em} George\\ \hspace{.5em} Bush} $\downarrow$} & \raisebox{1.3\height}{2003-05} & \raisebox{1.3\height}{2003-12} &\raisebox{1.3\height}{500}& \includegraphics[width=0.3\textwidth]{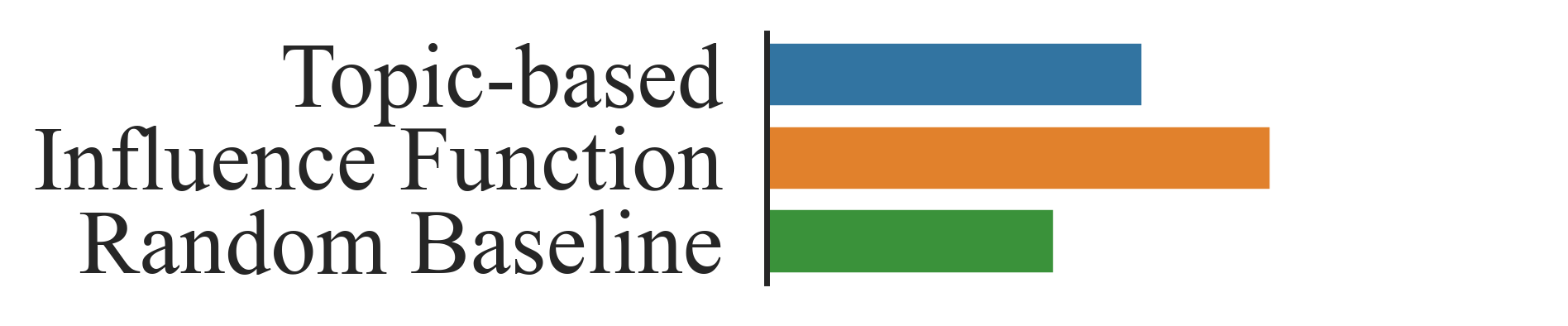} & \includegraphics[width=0.3\textwidth]{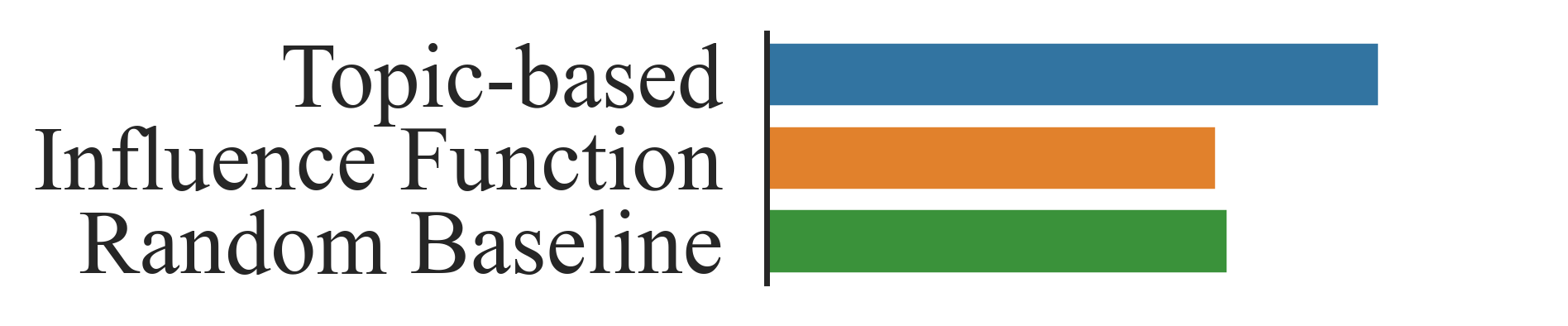}\\
         \hline
        \multicolumn{6}{c}{\makecell{The most salient topic words: capture, iraq, blair, foreign, saddam hussein}}
         \\ \hline
    \end{tabular}
    \caption{Textual source analyses for moral change toward entities in  historical events. Arrows show the directions of the moral polarity change. Column ``N'' shows the number of  articles retrieved in each time window. The  influence set size is $10\%$ of $N$. Bars under ``Influence comparison'' show inverse $\Delta{J}$s (lengthier for greater influence) under the three methods. ``Coherence comparison''  compares mean coherence ($E[H]$) of source text retrieved. The most salient words under the topic-based  method are provided.}
    \label{tab:nyt_all}
\end{table*}

{\bf Evaluation metrics.} 
We consider a baseline inspired by influence function~\citep{cook1980characterizations} to retrieve a set of  documents as the textual source of moral change. We compare this set to the documents retrieved by the topic-based source model based on the  metrics of 1) degree of influence and  2) coherence of the retrieved source.

To assess models based on degree of influence, we use $\Delta{J}$. We first generate a null distribution via perturbing the dataset. The dataset used here is a set of documents published in $t \sim t+\Delta{t}$ that mention entity $e$ (i.e., $D_{e, t\sim t+\Delta{t}}$). To construct the null distribution, we choose a random set of documents from $D_{e, t\sim t+\Delta{t}}$, denoted as $D^*$, and measure the influence of set $D^*$ on moral sentiment at a change point using $\Delta{J}$ in Equation~\ref{eq:4}. We repeat this process until we generate $10,000$ random document sets. The set of documents that minimizes $\Delta{J}$ significantly ($\alpha = 0.05$) compared to the null distribution would be the source text. These documents form a subset that provides the maximal perturbation to the moral sentiment estimated at the change point. For the topic-based model, we select the source set by choosing documents with the highest $p(topic = o | d)$, where $o$ is the topic minimizing Equation~\ref{eq:2}. 
The size of the source documents set for both the influence function and topic-based model would be $10\%$ of $|D_{e, t\sim t+\Delta{t}}|$. 
We then compare $\Delta{J}$ of these two sets. A lower value for $\Delta{J}$ indicates greater influence and hence a more effective identification of the source documents.

We define $E[H]$ to assess the coherence in the retrieved source documents. $E[H]$ is the average pairwise cosine similarity among a set of retrieved documents (i.e., news articles in this case). We consider  coherence a desirable property because the sources responsible for moral change toward an entity should ideally reflect a consistent set of content. The coherence metric evaluates whether  the retrieved source documents indeed form a consistent set of text. We use  Word2Vec embeddings to estimate the cosine similarities of news articles based on their headlines. Equation~\ref{efq:coherence} defines this metric for a document set $D$. In this equation, $V_{h_{d_i}}$ corresponds to the vector representation of the headline of news article $d_i$. For both the topic-based model and influence function, $E[H]$ is estimated on the same set of documents as for $\Delta{J}$:

\begin{equation}
    \begin{aligned}
    E[H] = 
    \frac{1}{|D|(|D| - 1)} 
    \sum_{d_i \in D} \sum_{\substack{d_j \in D\\ i \neq j}} \frac{V_{h_{d_i}}. V_{h_{d_j}}}{\|V_{h_{d_i}}\|\|V_{h_{d_j}}\|} &&&
    \end{aligned}
    \label{efq:coherence}
\end{equation}

We also consider a random baseline which arbitrarily retrieves the same number of documents as the topic-based and influence function methods.

Figure~\ref{fig:method_vis} illustrates $\Delta{J}$ and $E[H]$ based on NYT news about entity {\it Bill Clinton} from 1997-12 to 1998-03. The degree of influence and the coherence under the topic-based model are greater than those of the influence function and the random baseline. In particular, we observe that the topic retrieved as the source of the negative change in moral polarity of {\it Bill Clinton} is associated with the Clinton-Lewinsky Scandal (salient topic words include {\it lawyer}, {\it Starr}, {\it Lewinsky}, {\it Jones}), while the articles selected by influence function (salient words include {\it plan}, {\it political}, {\it senate}, {\it patience}) and the random baseline (sample words include {\it Iraq}, {\it Democrat}, {\it world}, {\it battle}) show minimal agreement in the context and no relevance to the ground-truth historical scandal of the period. The table in Appendix B shows the headlines of randomly sampled articles retrieved as sources of moral sentiment change by the three models.

For a more comprehensive evaluation, we select the following well-known historical events and entities: George H. W. Bush for Gulf War (1990-1991), Bill Clinton for the Clinton-Lewinsky Scandal (1997-1998), George W. Bush for September 11 attacks (2001), China for the SARS outbreak (2002-2004), George W. Bush and Saddam Hussein for the Iraq invasion (2003-2004). The time resolution for our analysis is by month. For each entity and event, we extract all the articles published in NYT that mention the entity at least once. We use the dynamic topic model to derive $10$ topics for each of the entities in the mentioned periods. We focus on assessing the models along the moral polarity dimension that has relatively clear-cut ground-truth for the historical incidents.

\begin{figure}[ht!]
\centering
\includegraphics[width=0.45\textwidth]{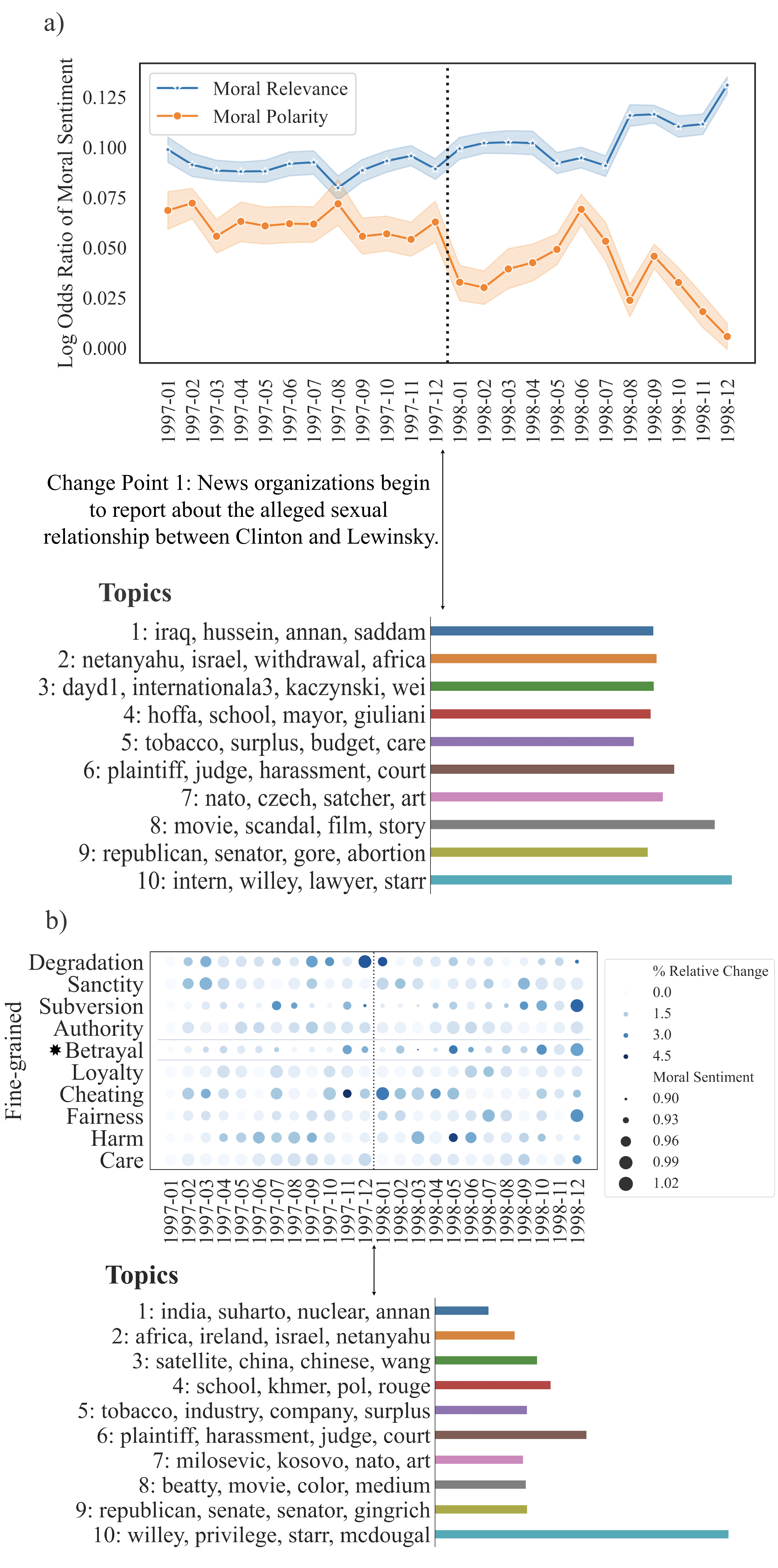}
\caption{Textual source analysis of moral  change toward {\it Bill Clinton} from 1997-1998. a) Changes in moral relevance and polarity. The vertical dashed line shows change point aligned with the start of the scandal. The bar plot below shows the topics and their relative contributions to the change. b) Fine-grained moral sentiment change toward {\it Bill Clinton}. The bottom plot shows the topical contributions in change along the Betrayal dimension from 1997-12 to 1998-04. Topic 10 is the most salient source.}
\label{fig:finalbillclinton}
\end{figure}

Table~\ref{tab:nyt_all} summarizes the result per entity and event. First, all the topics identified by the topic-based source model align with the (advent of) historical events. For instance, the negative change in the moral polarity toward  {\it George H. W. Bush} detected between 1990-07 and 1991-02 is associated with the topic of {\it Iraq} and {\it Saddam Hussein}. 
Comparisons on $\Delta{J}$ between the topic-based model and influence function indicate that these methods are equally effective in terms of the influence of the source documents ($p = 0.348$ via paired t-test), while the topic-based model significantly outperforms the random baseline ($p < 0.01$). Moreover, the topic-based model significantly outperforms the influence function and the random baseline ($p < 0.05$) in the expected coherence of the retrieved source documents (i.e., $E[H]$). This set of results shows that the topic-based model is on par with the established influence function retrieving influential source documents that underlie moral sentiment change, and it is significantly more effective in selecting coherent articles relevant to the source of moral  change. 
It is also important to note that although the influence function is designed to retrieve the most influential documents, it is computationally prohibitive to exhaustively search all possible documents, and here we applied a random search. In contrast, the topic-based model can also retrieve a set of influential documents, but it does not require an exhaustive iteration through all possible sets. Figure~\ref{fig:finalbillclinton} illustrates and interprets the  source analysis for the Clinton-Lewinsky Scandal.

\subsection{Application to textual source analysis of moral  change in COVID-19 news}
\label{sec:covid}

In the final case study, we apply our framework to textual source analyses of moral  change in COVID-19 news. Differing from the NYT case where we focused on evaluating moral changes against known historical events, here we focus on a real-time exploratory analysis of the COVID-19 news for four entities: Donald Trump, Anthony Fauci, Andrew Cuomo, and China.\footnote{We use a sliding window for change point analysis on a weekly scale, and each time point is a week starting from January, 2020.}

\begin{figure}[!ht]
\centering
\includegraphics[width=0.43\textwidth]{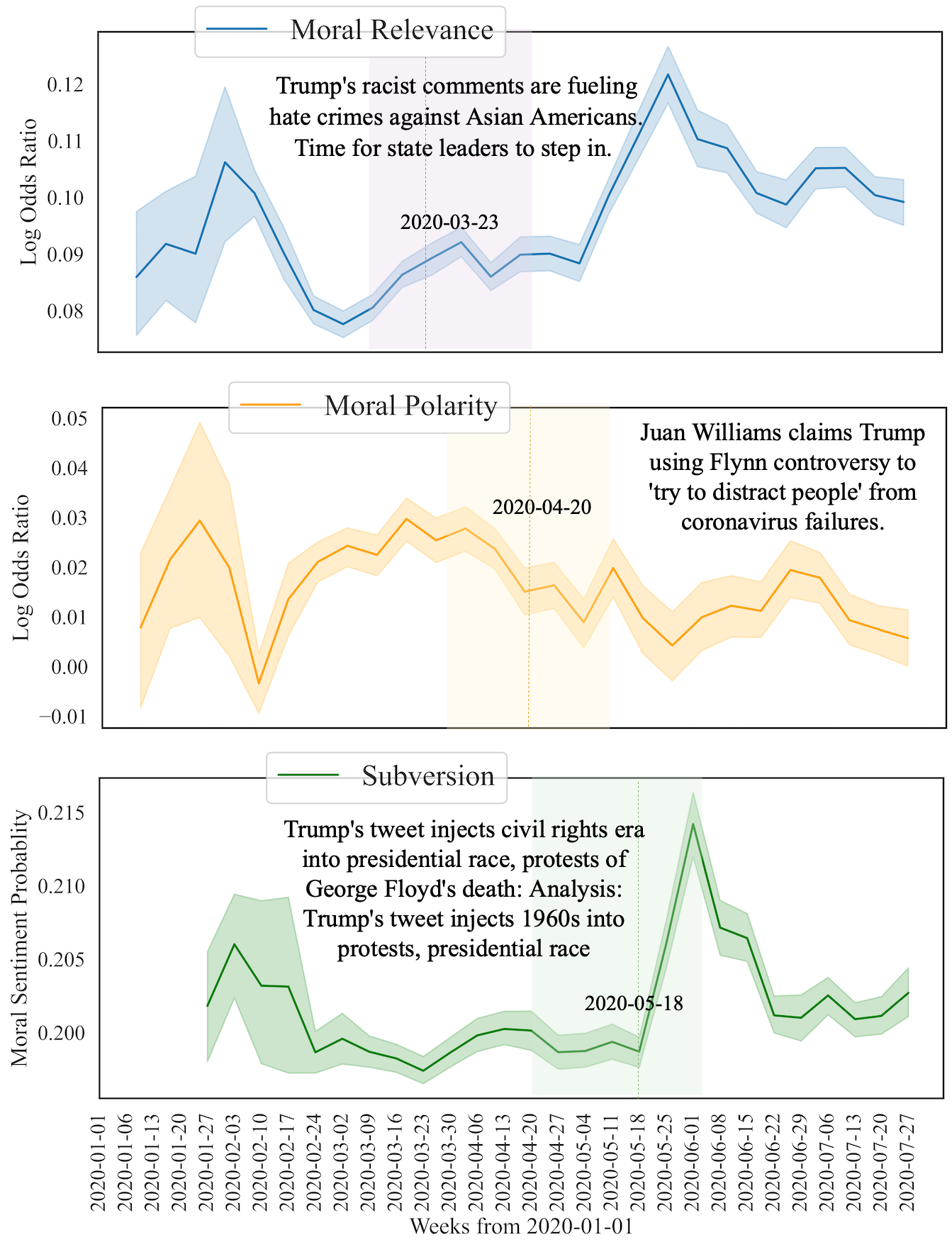}
\caption{Textual sources of moral change toward {\it Donald Trump} in COVID-19 news. Shaded boxes show the sliding windows where the change points are detected. The topic-based model finds the most salient topic, annotated with the headline of a high-probability article.}
\label{fig:trump}
\end{figure}

Figure~\ref{fig:trump} shows the moral source analysis of {\it Donald Trump}. The topics selected for each change point align well with the notable incidences in COVID-19 pandemic (as annotated), suggesting how such source events can be traced in short time windows from text. The top row shows the time course of moral relevance for {\it Donald Trump}. Some relevant topic words are {\it china},  {\it blame}, {\it disinformation}, and {\it asian}. The retrieved relevant words in the middle row, reflecting a moral polarity change are {\it flynn},  {\it ratcliffe},  {\it fbi}, and {\it investigation}. The bottom row shows similar analyses for subversion which is one of the 10 moral foundations. The changing point occurs at week of 2020-05-18, and the relevant terms include {\it george floyd}, {\it police}, {\it protest},  and {\it riot}. For a more in-depth analysis, we apply the topic-based source model to all four entities. The table in Appendix C summarizes the results. Certain events during the pandemic had significant impact on how the entities are morally portrayed in the news. For example, {\it George Floyd} incident is attributed to be the source of the increase in subversion for \emph{Donald Trump}. The shift happens after the week starting on May 18\textsuperscript{th}, which is close to the incident date May 25\textsuperscript{th}. Our model also identifies context when an entity becomes morally relevant. For example, it finds an increase in moral relevance for {\it Anthony Fauci} during June 22\textsuperscript{nd} to July 27\textsuperscript{th} with source topics concerning the conspiracy theories of COVID-19 treatment in social media, and the conflicts between Fauci and Trump.

\section{Discussion and conclusion}
We have presented an unsupervised framework that uses topical information to infer textual sources of moral  change for entities. Our work extends existing NLP methods for moral change inference by identifying the origins of moral change over time. With evaluations on a diverse set of data, we show that our method captures both the fine-grained human moral judgments and  coherent source text of moral change relevant to social events.

Our approach differs from work on general sentiment inference, partly because moral sentiment has a more fine-grained and hierarchical structure that involves inference at three different tiers, e.g., moral relevance, moral polarity (vice vs. virtue), and moral foundations. Previous work has evaluated rigorously models 
that capture this 3-tier moral hierarchy \cite{xie2020text}. Our framework builds on this study by characterizing the textual source of moral change at each of the three tiers. Although moral polarity can overlap with general sentiments such as good and bad, our framework captures moral sentiment beyond this dichotomous dimension. For instance, an increase in the moral relevance of an entity can be driven by an increase in moral
authority, which may or may not involve any positive or negative
sentiment (see Appendix C for examples). In this respect, moral sentiment captured by our framework
can be dissociated with sentiment portrayed in the traditional NLP literature.

Our work makes minimal claims about the causes of moral change. Our focus here is to identify salient topics as the source of moral sentiment change. This topical information can be a proxy to world events that trigger changes in moral perception toward an entity. Identifying the causes of moral change beyond textual sources studied here can be an exciting yet challenging direction.

Our framework also offers opportunities for further exploration of entity-based moral sentiment change. Future work may explore how different  media platforms vary in the moral sentiments that they convey towards entities (e.g., public figures) and the sources of this variation.

\section*{Acknowledgements}

We thank Graeme Hirst, Jennifer Stellar, Lea Frermann, and Matthew Fienberg for their feedback on our manuscript. AR is funded partly by a Schwartz Reisman Institute for Technology and Society Graduate Fellowship. This work was supported by a NSERC Discovery Grant RGPIN-2018-05872, a SSHRC Insight Grant \#435190272, and an Ontario ERA Award to YX.

\section*{Broader impact and ethical statement}

Our study applies natural language processing to the source identification of moral change triggered by high-impact historical and contemporary social events. The framework we have developed provides an automated and scalable tool for interdisciplinary scholars including computational linguists, psychologists, and social scientists to quantitatively investigate the origins of moral change.

Some of our analyses on social media and news articles can be considered sensitive. The results that we report here do not represent our personal beliefs or opinions and could potentially be influenced by the biases contained in the datasets.


\bibliographystyle{acl_natbib}
\bibliography{anthology,emnlp2021}

\clearpage

\appendix

\section{Details of data pre-processing}

We take the following steps in pre-processing for a document and a query entity: 1) We find all the mentions of the entity in the document using the co-reference resolution module specified in the paper. 2)  We discard all the sentences in the document that do not include any mention of the entity. 3) We lemmatize the tokens in the documents using the \texttt{spaCy} English model. 4) We remove the entity with all its mentions, function words, and all the other words if they are classified as morally irrelevant using the centroid model following  \citet{xie2020text}. 5) We take an average of the word embeddings of the remaining tokens to derive the feature vector of the document. If the task is based on the BERT embeddings, we skip step 4, but to derive the vector representation of the document, we take an average of the tokens that pass through step 4 in the final layer.

In \citet{xie2020text}, the centroid model compares the Euclidean distance of an embedded input to the center of morally relevant words and the center of morally neutral words. The distances are then transformed to probabilities using a softmax function.

Our framework models moral sentiment as a hierarchical concept under the three tiers of 1) moral relevance, 2) moral polarity, and 3) 10 fine-grained moral foundation categories. All the calculations for the moral sentiment dimensions are performed on the documents that satisfy this hierarchical framework. For instance, when estimating the moral polarity, the documents classified as morally irrelevant are discarded. Similarly, for each fine-grained category, we discard documents with an opposing moral polarity.

\clearpage

\section{Comparison of source news articles retrieved for Bill Clinton case study}
\begin{table}[htbp!]
\centering
\begin{tabular}{ll}
\hline
\textbf{Headlines of news articles retrieved by topic-based source model} & \\
\multirow{2}{*}{Lawyers for Jones Get More Response Time}&\\
\multirow{2}{*}{Whispered Secrets Start a Loud Debate}&\\
\multirow{2}{*}{Starr Is Right to Question White House Aide; Having It Both Ways}&\\
\multirow{2}{*}{Lewinsky's Bookstore Purchases Are Now Subject of a Subpoena}& \\\\
\hline
\textbf{Headlines of news articles retrieved by influence function method} &  \\
\multirow{2}{*}{A Waggish Tale In Washington...}&\\
\multirow{2}{*}{Starr Subpoenas Notes and Case Files of Lewinsky's Former Lawyer}&\\
\multirow{2}{*}{Would Punishing Iraq Carry Too High a Price? Vietnam's Lesson}&\\
\multirow{2}{*}{Day of Facing the Nation, Meeting the Press, Etc.}&\\
\\
\hline
\textbf {Headlines of news articles retrieved by random baseline} &{}\\
\multirow{2}{*}{Public Radio Hosts Drop In and Maybe Stay Too Long} & \\
\multirow{2}{*}{Book Agent Advised Taping Accusations} & \\
\multirow{2}{*}{After Derailing Trade Bill, Labor Sets Ambitious Goals} & \\
\multirow{2}{*}{Yes, a Surplus Would Help, But Tough Choices Remain} & \\
\\
\hline
\end{tabular}
\caption{Headlines of 4 randomly sampled news articles retrieved by the three models
as source for moral sentiment change toward \textit{Bill Clinton} during Clinton-Lewinsky Scandal (1997-12 to 1998-03). }
\label{tab:models_articles}
\end{table}

\clearpage
\section{Additional results from moral change source analysis for entities in COVID-19 news}
\label{sec:appendix-a}
\hfuzz=0.64pt
\begin{table*}[b!]
    \begin{tabular}{l p{2cm} p{2cm} p{2cm} c}
        \textbf{Entity} &  \textbf{Initial point} & \textbf{Ending point} &  \textbf{Moral Dimension} & \textbf{Topic Words}
        \\
        \Xhline{2pt}
        \\
        Trump & 2020-03-23 & 2020-04-20 & Relevance $\uparrow$ &
        \makecell{conspiracy, xenophobic, disinformation\\ china,  originate, blame, asian} \\
        {} & 2020-05-18 & 2020-06-22 & Relevance $\uparrow$ &
        \makecell{juneteenth, police, black, floyd, racism, \\
        brutality, protest, racial, minneapolis} \\
        {}  & 2020-04-20 & 2020-05-11 & Polarity $\downarrow$ & 
        \makecell{flynn, muir, mcenany, miller, obama\\ collusion, ratcliffe, whistleblower }\\
        {}  & 2020-05-18 &	2020-06-01 & Subversion $\uparrow$ &
        \makecell{killing, george floyd, protest, black\\minneapolis, peaceful, racism, murder} \\
        \hline
        Fauci  & 2020-06-22 & 2020-07-27 & Relevance $\uparrow$ &
        \makecell{sinclair, twitter, mikovit, conspiracy \\facebook, vaccine, mask, video} \\
        {}  & 2020-06-29 & 2020-07-20 & Fairness $\uparrow$ &
        \makecell{disapprove, statue, cain, goya, gop\\ electoral, biden, campaign, tulsa}  \\
        \hline
         Cuomo  & 2020-05-11 & 2020-06-01 &	Relevance $\uparrow$ &
        \makecell{george floyd, cop, demonstration \\injustice, black, peaceful, protest, racism} \\
        {}  & 2020-03-30 & 2020-05-04 & Polarity $\downarrow$ & 
        \makecell{14-day, death, flatten, epicenter\\lockdown, peak, social distancing, reopen} \\
        {}   & 2020-05-25  & 2020-06-22 & Polarity $\downarrow$&
        \makecell{george floyd, cop, demonstration \\injustice, black, peaceful, protest, racism} \\
        {}  & 2020-03-23 &	2020-04-13 & Cheating $\downarrow$ &
        \makecell{14-day, death, flatten, epicenter\\ lockdown, peak, social distancing, reopen} \\
        \hline
        China  & 2020-03-23	& 2020-04-20 & Relevance $\uparrow$ & 
        \makecell{blame, disinformation,
         trump, conspiracy\\accountable, downplay} \\
        {}  & 2020-05-11	& 2020-05-25 & Relevance $\uparrow$ & 
        \makecell{hong kong, freedom, democracy,\\ economic, tension, territory} \\
        {}  &  2020-02-24 &  2020-03-23 & Polarity $\uparrow$ & 
        \makecell{iran, ban, passenger, flight \\ quarantine, cruise, case, korea, japan} \\
        {}  & 2020-05-11	& 2020-05-25 & Polarity	$\uparrow$ &
        \makecell{hong kong, freedom, democracy\\ economic, tension, territory} \\
        {}   & 	2020-02-24 & 2020-03-23	& Authority  $\uparrow$ &
        \makecell{flu, disease, sick, test, care, influenza \\ cough, respiratory, ventilator, covid-19} \\
    \end{tabular}
           
    \caption{Source analyses of moral sentiment change of entities in COVID-19 along different moral dimensions. Arrows indicate the polarities of change. The most salient words from the source topics are shown for each entity.}
    \label{tab:covid_all}
\end{table*}

\end{document}